\documentclass[10pt,a4paper,twocolumn]{scrartcl}
\usepackage[utf8]{inputenc}
\usepackage{geometry}
\usepackage{glossaries}
\usepackage{amsmath,amssymb}
\usepackage[detect-all,binary-units]{siunitx}
\usepackage{bm}

\usepackage{subcaption}
\usepackage{authblk}
\usepackage{hyperref}
\usepackage{cleveref}
\usepackage{ifthen}
\usepackage{graphicx}
\usepackage[english]{babel}
\graphicspath{{figures/}{figures/inkscape/}{figures/py/plots/}}
\def\BSS2{\mbox{BrainScaleS-2}}

\usepackage{tikz}

\definecolor{blue}{HTML}{56CCDB}%
\definecolor{red}{HTML}{FF5656}%
\definecolor{green}{HTML}{60EA90}%
\definecolor{yellow}{HTML}{FFFC31}%

\DeclareSIUnit\steps{steps}

\newif\ifarxiv
\arxivfalse

\setkomafont{title}{\normalfont\Huge}
\setkomafont{section}{\normalfont\Large}
\setkomafont{subsection}{\normalfont\large}

\newacronym{ppu}{PPU}{plasticity processing unit}
\newacronym{lif}{LIF}{leaky integrate-and-fire}
\newacronym{fpga}{FPGA}{field-programmable gate array}
\newacronym{lstm}{LSTM}{long short-term memory}
\newacronym{simd}{SIMD}{single instruction, multiple data}
\newacronym{cpu}{CPU}{central processing unit}

\newacronym{tn}{TN}{\textbf{t}angential \textbf{n}oduli}
\newacronym{tl}{TL}{\textbf{t}angential \textbf{l}ower division of the central body}
\newacronym{cl1}{CL1}{\textbf{c}olumnar \textbf{l}ower division of the central body, type 1}
\newacronym{tb1}{TB1}{\textbf{t}angential protocerebral \textbf{b}ridge, type 1}
\newacronym{cpu1}{CPU1}{\textit{\textbf{c}olumnar \textbf{p}rotocerebral bridge/\textbf{u}pper division of the central body, type 1}}
\newacronym{cpu4}{CPU4}{\textit{\textbf{c}olumnar \textbf{p}rotocerebral bridge/\textbf{u}pper division of the central body, type 4}}
\newacronym{m}{M}{motor}

\usepackage[numbers,sort&compress]{natbib}
\title{Emulating insect brains for neuromorphic navigation}

\date{\today}

\begin{document}

\author[1,3]{Korbinian Schreiber} %
\author[1]{Timo Wunderlich} %
\author[1]{Philipp Spilger} %
\author[1]{Sebastian Billaudelle} %
\author[1]{Benjamin Cramer} %
\author[1]{Yannik Stradmann} %
\author[1,4]{Christian Pehle} %
\author[1]{Eric M{\"u}ller} %
\author[2]{Mihai A. Petrovici} %
\author[$\S$,1]{\\Johannes Schemmel}
\author[$\S$,1,$\dagger$]{Karlheinz Meier}

\affil[$\S$]{Shared senior authorship}
\affil[$\dagger$]{Deceased}
\affil[1]{Kirchhoff-Institute for Physics, Heidelberg University}
\affil[2]{Department of Physiology, University of Bern}
\affil[3]{Institut für Innovation und Technik (iit), Berlin}
\affil[4]{Cold Spring Harbor Laboratory}

\maketitle

\begin{abstract}
\noindent\bfseries
Bees display the remarkable ability to return home in a straight line after meandering excursions to their environment.
Neurobiological imaging studies have revealed that this capability emerges from a path integration mechanism implemented within the insect's brain.
In the present work, we emulate this neural network on the neuromorphic mixed-signal processor \BSS2 to guide bees, virtually embodied on a digital co-processor, back to their home location after randomly exploring their environment. 
To realize the underlying neural integrators, we introduce single-neuron spike-based short-term memory cells with axo-axonic synapses.
All entities, including environment, sensory organs, brain, actuators, and the virtual body, run autonomously on a single \BSS2 microchip.
The functioning network is fine-tuned for better precision and reliability through an evolution strategy. 
As \BSS2 emulates neural processes 1000 times faster than biology, 4800 consecutive bee journeys distributed over 320 generations occur within only half an hour on a single neuromorphic core.
\end{abstract}
\vspace{0.5em}

{
\bfseries
\itshape
\noindent
path integration, navigation, central complex, bee, short-term memory, BrainScaleS, mixed-signal, neuromorphic, spiking, neural networks

}

\glsresetall

\section{Introduction}
Insects are capable of navigating complex real-world environments with remarkable efficiency, speed and robustness.
Besides the remarkable adaptation of their bodies for this task this is due to their neural anatomy which evolved over a period of at least 520 million years~\citep{ma2012complex}.
Being capable of mastering numerous difficult tasks such as homing~\citep{collett2002memory}, migrating over distances of more than 1000 kilometers~\citep{chapman2002high, adden2020brain}, remembering and communicating spatial locations~\citep{von1967dance}, or learning visual cues~\citep{nityananda2014can}, even large insect brains measure only a few cubic millimeters in volume and often consist of less than a million neurons~\citep{burrows1996neurobiology, menzel2001cognitive}.

Within the last decade, research in experimental neuroscience has brought about a completely new quality in the measurement of neuronal structures and processes, providing groundbreaking insights into their working principles.
Advances in biological imaging and data processing resulted in almost complete connectomes of insect brains~\citep{chiang2011three,takemura2013visual,takemura2017connectome,xu2020connectome}, while methodological combinations of genetic modification, bio-imaging, and virtual reality for fixated animals made it possible to record real-time cell-level neural activity of brains engaged in realistic interactions with the environment~\citep{stone2017anatomically,huang2020virtual}.
Whereas earlier lesion-based studies have shown that insect brain structures like the \textit{central complex}~\citep{turner2016insect} are somehow involved in orientation and navigation~\citep{neuser2008analysis}, modern calcium-based imaging techniques~\citep{grienberger2012imaging} provide detailed time-resolved insight.
For example, imaging the head-fixed fruit fly \textit{Drosophila  melanogaster} \textit{in vivo} in virtual environments has revealed that subcomponents of the \textit{central complex} track azimuthal visual cues~\citep{seelig2015neural} and indicate and integrate the fly's heading direction~\citep{green2017neural}.
Similar experiments on the sweat bee \textit{Megalopta genalis} have even revealed enough insight to construct a fully functional physiological model for path integration-based navigation~\citep{stone2017anatomically}, deciphering a mechanism that lets the insect keep track of its home location by integrating over its current heading velocities.

While biological experiments continue to unveil intricate physiological details of these tiny nervous systems, studying such abstract neural network models in simulations can confirm and sharpen our understanding of how nervous structures correspond to their respective functions~\citep{stone2017anatomically, kagioulis2020insect, zhao2021predictive}. 
As those functions are often defined by interactions of the entire animal with its environment, faithful and valid simulations need to encompass also components beyond the neurons.
This might include the environment, the animal's body, its sensory and motoric organs, or the information de- and encoders translating between external physics and nervous signals. 
All those components can be implemented virtually in simulations~\citep{falotico2017connecting, feldotto2022deploying, feldotto2022neurorobotics, bennett2021learning} or physically through robots~\citep{dupeyroux2019antbot, husbands2021recent}.

Similarly, the dynamics of the attached neural networks can be solved numerically within simulations~\citep{markram2015reconstruction,jordan2018extremely,knight2018gpus,sarma2018openworm}, or emulated using physical, neuromorphic circuits~\citep{kreiser2018pose, wunderlich2019demonstratingstudy, kreiser2019error, kreiser2019self, kreiser2020chip, hajizada2022interactive, sandamirskaya2022neuromorphic, stradmann2023biomorphic}.
Those circuits are usually densely integrated on CMOS-based neuromorphic microchips and mimic the architecture and properties of biological neural networks.
Unlike common \glspl{cpu}, which typically operate on data within a single large block of memory, neuromorphic processors tend to perform many parallel computations using multiple processing units and distributed memory. 
While some implementations are based on specialized digital building blocks \citep{furber2014spinnaker, merolla2014million, davies2018loihi, frenkel20180, frenkel2019morphic, orchard2021efficient}, others recreate the dynamics of physiological membrane potentials in terms of analog voltages and currents, thus, emulating physical models of the biological archetypes \citep{benjamin2014neurogrid, qiao2015reconfigurable, moradi2017scalable, neckar2018braindrop, schemmel2017accelerated, pehle2022brainscales2, le202364}.
Also in the field of analog neuromorphic engineering, rapid advances and developments have taken place especially within the last decade, revealing unprecedented versatility at outstanding power efficiencies~\citep{wunderlich2019demonstratingstudy, wunderlich2019b, billaudelle2020versatile, billaudelle2021structural, kungl2019accelerated, schreiber2020closed, schreiber2020accelerated, goeltz2021fastanddeep, cramer2022surrogate}.

In the present work we merge recent seminal discoveries in experimental neuroscience with a number of recent innovations in neuromorphic engineering into a cohesive neurorobotic agent.
Relying on an accelerated analog neuromorphic nervous system, we reproduce the ability of bees to return to their nest's location after excursions through a two-dimensional environment.
While previous case studies have successfully mapped various subcomponents of the networks used by insects or mammals for path integration onto neuromorphic substrates~\cite{massoud2010neuromorphic,kreiser2018pose,kreiser2018neuromorphic,kreiser2019self,kreiser2020chip}, we demonstrate a fully embodied and autonomous neuromorphic agent within a single \BSS2 neuromophic prototype chip~\citep{aamir2018accelerated,pehle2022brainscales2,schreiber2020accelerated,billaudelle2020versatile,schreiber2020closed}.
The agent's body and the environment are simulated on a digital on-chip coprocessor, while the neural network is physically emulated on the chip's analog neuromorphic core with sensory inputs and motor outputs connecting the two domains via mixed-signal circuits.
Due to the intrinsic time scales of the analog circuits, the neural dynamics proceed 1000$\times$ faster than in biology.
This speed-up is independent of the network size and allows for rapid prototyping during the initial design process, when architectural decisions are still to be made or when parallelizing stochastic parameter evaluations is not feasible.
The acceleration also benefits optimization algorithms which improve the network's performance by successive incremental parameter updates~\citep{wunderlich2019demonstratingstudy,wunderlich2019b,billaudelle2020versatile,billaudelle2021structural,kungl2019accelerated,goeltz2021fastanddeep,pehle2022brainscales2,cramer2022surrogate}.
We demonstrate this by employing an evolution strategy algorithm for fine-tuning the agent's performance through synaptic weight updates.
As an example, spawning 320 successive generations of 15 neuromorphic insects and evaluating the navigation capabilities of each individual during \SI{3.3}{\minute} flights takes only around \SI{0.5}{\hour} on a single neuromorphic processor without parallelization.
Executed in real time, the flight simulations alone would take more than 11 days.

Rendering the foundational neural network model derived from \textit{Megalopta genalis}~\citep{stone2017anatomically} compatible with the neuromorphic circuits on \BSS2 additionally required some essential adaptations and adjustments.
This included, most importantly, the transition from abstract, analytical firing rate neurons to spike-based \gls{lif} neurons.
Besides common firing rate models which map summed input activities through a nonlinear response function to an output activity, the original model crucially relies on heuristically defined equations responsible for storing and integrating the activities of presynaptic neurons.
To biologically plausibly implement a functionally identical mechanism based on spiking \gls{lif} neurons, we introduce a single-cell short-term memory mechanism based on axo-axonic synapses.
While the resulting model has been developed for \BSS2, it is flexible enough to be ported to other neuromorphic systems as well.
For example, sub-threshold neuromorphic platforms, like ROLLS~\citep{qiao2015reconfigurable} or DYNAPs~\citep{moradi2017scalable}, could adapt the network to facilitate ultra-low-power robotic control units.
Other applications requiring more robustness and repeatability could instead make use of digital neuromorphic systems like SpiNNaker~\citep{furber2014spinnaker,mayr2019spinnaker}, Intel's Loihi~\citep{davies2018loihi,orchard2021efficient} or IBM's TrueNorth~\citep{merolla2014million}.
Being compatible with a wide range of spiking neural models, the presented model can also serve as a building block in larger insect-inspired systems or in neuro-inspired control units.

\section{Results}

\subsection{\BSS2}
\BSS2 is a mixed-signal neuromorphic computing system~\citep{aamir2018accelerated,pehle2022brainscales2}.
The second-generation prototype chip used in this work contains an analog neuromorphic core with 32 physical neurons~\citep{aamir2018accelerated} with 32 synapses each.\footnote{The full-scale \BSS2 chip has 512 neurons with 256 synapses each.}
The neuron dynamics follow the \gls{lif} equation $C_{\mathrm{m}}\dot{V}_{\mathrm{m}} = -g_{\mathrm{l}}(V_{\mathrm{m}} - V_{\mathrm{l}}) + I_{\mathrm{syn}}$, where $C_{\mathrm{m}}$ is the membrane capacitance, $V_{\mathrm{m}}$ the dynamic membrane voltage, $g_{\mathrm{l}}$ the leak conductance, $V_{\mathrm{l}}$ the static leak or resting potential, and $I_{\mathrm{syn}}$ the dynamic synaptic input current.
Whenever the membrane potential crosses the threshold voltage $V_{\mathrm{th}}$, it is reset to $V_{\mathrm{reset}}$ and the neuron's spike counter is incremented.
The post-synaptic spike is represented as a digital data package that carries the unique address of the firing neuron.
It is sent off-chip to an \gls{fpga} responsible for recurrent spike routing (later versions of the platform include an on-chip routing engine) and turned into a pre-synaptic spike with a newly assigned target label.

Pre-synaptic spikes arrive at the mixed-signal synapse circuits where their target label is compared against the programmable synaptic label.
If the labels match, the circuit triggers a synaptic event (i.e., a current pulse) with an amplitude proportional to a \SI{6}{\bit} programmable synaptic weight $w$.
The resulting current pulse creates a post-synaptic potential on the dendritic capacitance which is finally transferred to the membrane capacitance of the receiving neuron.
Synapses can be either excitatory or inhibitory, resulting in a positive or negative postsynaptic potential, respectively.

The prototype chip includes an embedded digital processor clocked at \SI{100}{\mega\hertz}\footnote{On the full-scale \BSS2 chip, this processor is clocked at \SI{250}{\mega\hertz}.} that can access all on-chip components \citep{friedmann2016demonstrating}.
Its \gls{simd} vector unit can be used to efficiently read and write synaptic addresses and weights.
Since the processor's primary purpose is to compute synaptic weight updates, it is called the \gls{ppu}.
In this work, we employ the \gls{ppu} to simulate the agent and its environment, to inject spikes into the neural network from virtual spike sources, to manage the weight dynamics of the integration mechanism, for experiment control and for recording simulation states.

\begin{figure}
    \resizebox{\linewidth}{!}{\input{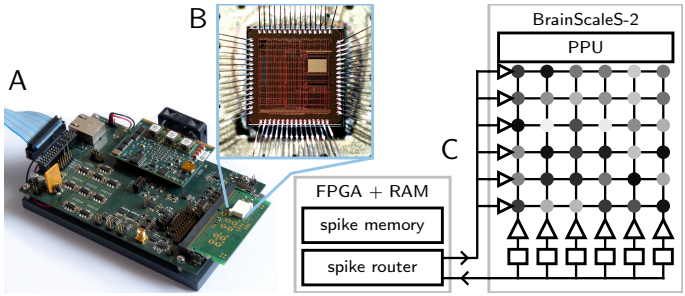}}
    \caption{\BSS2 prototype system.
    A) Entire system with \gls{fpga}, \BSS2 chip and periphery.
    B) \BSS2 prototype chip of the second generation.
    C) System schematic depicting the \gls{fpga} and the chip.
    Events coming from the \gls{fpga} are distributed to the synapse rows.
    The synaptic columns are connected to the triangular-shaped neurons that send spikes to the spike counters at the bottom and then back to the \gls{fpga}.}
    \label{fig:setup}
\end{figure}

\subsection{Network model and embodiment}
\label{sec:results_network}
The presented network model is based on \citet{stone2017anatomically} where a biologically plausible mechanism for path integration in bees is proposed.
We refer the reader to the referenced publication for an exposition of the biological background.
This section explains all functional aspects of the model and describes details about the presented implementation.

\begin{figure}
    \includegraphics[width=\linewidth]{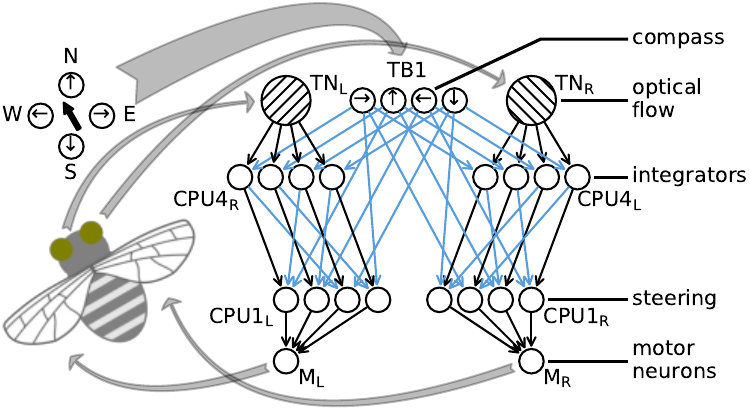}
    \caption{Network architecture.
    The black and blue arrows correspond to excitatory and inhibitory connections, respectively.
    The \gls{cpu1} and \gls{cpu4} populations are each divided into a left and right subpopulation, resulting in two logical hemispheres.
    The network topology relies on a physiological model proposed in \cite{stone2017anatomically}.}
    \label{fig:network}
\end{figure}

The state of the virtual insect is represented by four variables: the two-dimensional spatial position $(x, y)$, the head direction $\phi$ and the velocity $v$.
Each experiment starts with a spread-out phase lasting for \SI{50}{\milli\second} (equivalent to \SI{50}{\second} in the biological time frame) in which the agent is forced on a random walk.
This simulates a flight trajectory that the insect might take on its search for a food source.
During this time, the network receives sensory input from the head orientation and the optical flow across its eyes, but is disconnected from the motor units and does, thus, not control the insect's motion.

The neuronal representation of head orientation is modeled after the behavior of neurons in the \textit{central complex}, a part of the insect brain that is critically involved in navigation.
In particular, it is known that the \gls{tb1} group of neurons contained therein directly encodes the current head orientation like an internal biological compass.
Each \gls{tb1} neuron corresponds to a specific absolute direction and its activity is proportional to the angular proximity of the current head orientation with respect to this direction~\citep{heinze2007maplike, heinze2011sun, green2017neural}.
For example, when the insect flies straight north, the neuron encoding north will be most active, whereas the neuron encoding south will be least active (see also \cref{equ:tb1}).
While the insect's nervous system typically operates with a number of eight to nine such neurons, the resource constraints of the small \BSS2 prototype require reducing this number to four.
Each of the modeled \gls{tb1} neurons therefore corresponds to one of the four cardinal directions (see \cref{fig:network}).

Odometric information, i.e. information on the agent's speed, is derived from the optical flow across the left and right eye, respectively.
Also in this case, the neuronal encoding is modelled after physiological observations.
In particular, two \gls{tn} neurons, which penetrate the \textit{central complex}, coming from the \textit{lateral accessory lobes}~\citep{stone2017anatomically} spike with a rate proportional to the optical flow across the respective eye in one particular direction.
For example, if the insect flies forward with maximum speed, the firing rate of these cells increases to a maximum.
If it flies backward with sufficient speed, these neurons stop spiking.
A detailed mathematical description of all sensors is given in \cref{sec:methods_inputs}.

The output of both sensor populations is then processed by integrator neurons that mimic the \gls{cpu4} population found in the insect brain.
Those cells are grouped into two symmetrical subpopulations, each containing eight to nine neurons which are, like in \gls{tb1}, associated with the aforementioned azimuthal directions.
Again, this number is reduced to four cardinal directions due to resource constraints on \BSS2, resulting in a total of eight emulated \gls{cpu4} neurons.
In \citep{stone2017anatomically}, each \gls{cpu4} cell implements a heuristically defined differential equation that integrates the presynaptic activities of \gls{tn} and \gls{tb1} over time and outputs a numerical value in proportion to the accumulated value (see \cref{eq:cpu4_internal_state} in \cref{sec:neuron_model}).
As spiking \gls{lif} neurons are not capable of integrating or storing synaptic input signals in such a manner, we based the integration mechanism on axo-axonic synapses~\citep{kandel2000principles}. 
This synapse type plays an important role in the gill and siphon withdrawal reflex of the sea snail \textit{Aplysia} \citep{kandel1976cellular} but is commonly found in insect brains as well~\citep{schneider2016quantitative}.
Unlike in the more commonly found axodendritic synapses, the presynaptic neuron does not attach to the postsynaptic dendritic tree but to a postsynaptic axon terminal.
Here, its activity can modulate the weight of the following synapse~\citep{byrne1996presynaptic, lodish2012molecular}, effectively implementing an activity-driven synaptic integration mechanism (see \cref{eq:weight_update} in \cref{sec:neuron_model}).
Based hereon, the \gls{cpu4} population as a whole implements a working memory, in which the direction and distance to the home location is stored and continuously updated. 
Specifically, the vector connecting the current position to the origin is equal to the vector encoded by the \gls{cpu4} activities. 

The vector pointing toward the home location, as encoded in the output of the \gls{cpu4} population, is then compared to the current movement direction provided by the \gls{tb1} cells.
On \BSS2, this computation is performed by a symmetrically subdivided population of eight neurons in total modeled after biological \gls{cpu1} cells.
Each \gls{cpu1} subpopulation receives excitatory one-to-one input from \gls{cpu4} with identical source and target index, but also inhibitory one-to-one input where the index assignment is rotated by \ang{180} (see \cref{fig:network}).
This differential mapping helps suppress the common average activity of the entire \gls{cpu4} population while amplifying the activity differences between the individual pairs of opposing cells, emphasizing the stored direction.
Furthermore, the current direction of motion indicated by \gls{tb1} acts inhibitorily and rotated by \ang{+90} or \ang{-90} on the left or right \gls{cpu1} subpopulation.
Consequently, the summed activity of the left and right \gls{cpu1} subpopulation is inversely proportional to the alignment between the home vector and the current heading direction tilted to the right and left, respectively.
Hence, steering signals can be derived from \gls{cpu1} that guide the insect back to its nest.
The corresponding summations are carried out by a left and a right motor neuron $\mathrm{M_{L/R}}$.
These neurons were not derived from observations of the bee brain but introduced to transfer computational load from the digital processor on \BSS2 to the accelerated neuromorphic substrate, improving the efficiency of the entire emulation.

In the second part of the experiment, the return phase, the insect's motion is no longer externally imposed but derived from the outputs of these two motor neurons.
They influence the insect's motion by providing propulsion on the left and right hand side, similar to a tank drive.

All experiments run for $\SI{200}{\milli\second}$ and the insect is set to return after $t_{\mathrm{return}} = \SI{50}{\milli\second}$.
Translated to the biological time domain, this corresponds to a total duration of around \SI{3.3}{\minute} while the outbound journey lasts for \SI{50}{\second}.

Note that this duration is not intrinsic to model or hardware but can be set arbitrarily.
Here, we adjusted it to provide quick experimental throughput, while keeping the time scales of the neural processes and the time scales of the foraging in a biologically realistic relation to each other.
Specifically, the neuron dynamics take place within $\mathcal{O}(\SI{}{\milli\second})$ ($\mathcal{O}(\SI{}{\micro\second})$ on hardware) while the foraging happens within $\mathcal{O}(\SI{}{\minute})$ ($\mathcal{O}(\SI{100}{\milli\second})$ on hardware).

Moreover, as the memory resources on the \BSS2 prototype are limited, the chosen duration provides a decent tradeoff between time resolution and length of the recorded movements and neural activities.

By bringing together all the mentioned components and settings, the agent can be put into operation.

\subsection{Experiment execution}
\begin{figure}
    \includegraphics[width=\linewidth]{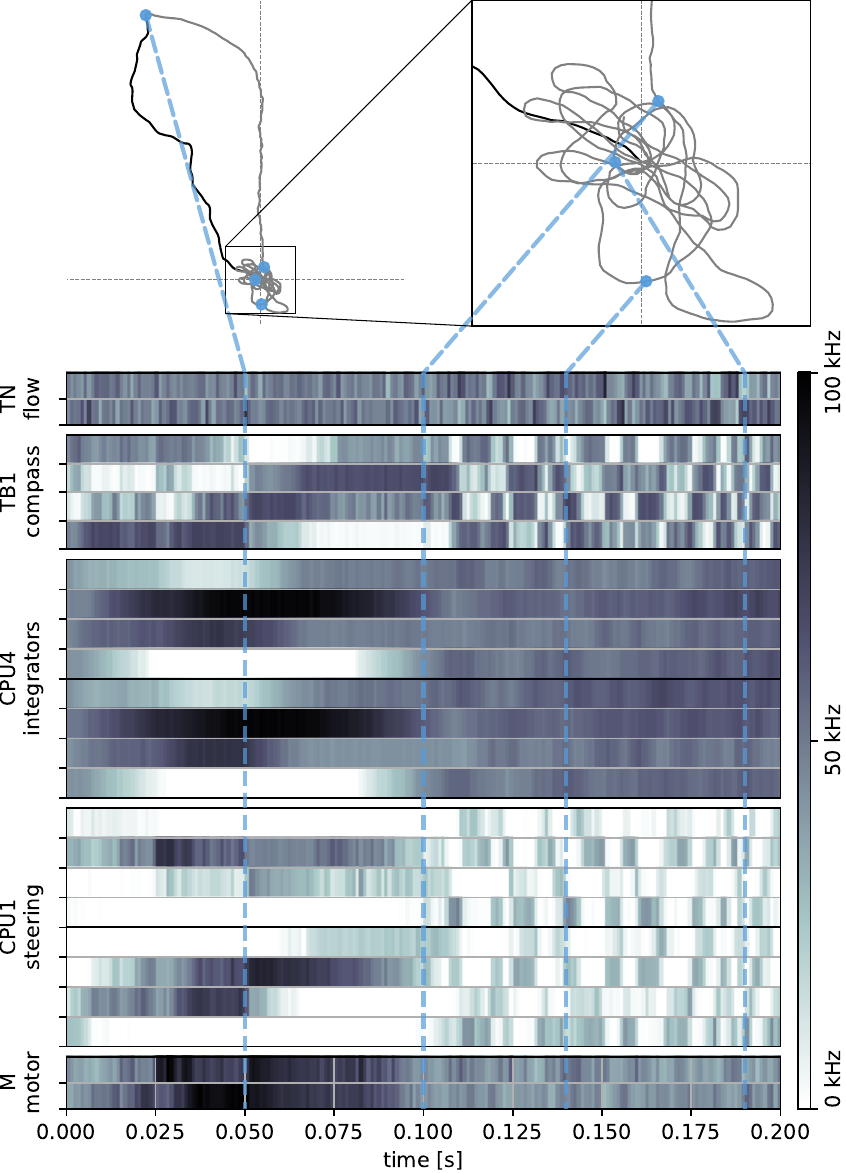}
    \caption{
        Network activity and trajectory.
        The top left plot shows the entire trajectory consisting of the outbound (black) and return phase (gray).
        The top right plot shows a zoom into a region measuring 5000 $\times$ 5000 steps around the origin where the agent loops around after his return.
        Four particular moments are highlighted as blue dots and connected to the network activity plot below: the moment of return, reaching the home location, and two situations during the looping phase.
        The neuron activities are measured as output spike counts in \SI{1}{\milli\second} intervals (corresponding to \SI{1}{\second} intervals in the biological time domain) and are displayed color-coded over time.
        The maximum count is approximately 100 which corresponds to an instantaneous spike rate of \SI{100}{\kilo\Hz} (equivalent to \SI{100}{\Hz} when translated into biological time scales).
    }
    \label{fig:spike_rates}
\end{figure}

\Cref{fig:spike_rates} shows the trajectory and network activity for a single journey.
While $t < t_{\mathrm{return}}$, the insect performs a random walk through its environment.
During this phase, it continuously integrates its current course, indicated by the compass neurons in \gls{tb1} and the odometric sensors \gls{tn}, within the left and right \gls{cpu4} populations.
While the activity among these neurons is initially homogeneous (at $t=\SI{0}{\second}$), the individual firing rates diverge as the insect moves away from its home location.
The firing rates of the \gls{cpu4} neurons collectively encode the current vector from the insect to the home location.

Simultaneously, the \gls{cpu1} populations compute the differences between this vector and the current left and right-tilted heading direction while the motor neurons $\mathrm{M_{L/R}}$ aggregate their activities.

After a fixed amount of time has passed (at $t=t_{\mathrm{return}}$), the motor neurons $\mathrm{M_{L/R}}$ assume control over the insect's motion and steer it back home. 
A discrepancy in activity between $\mathrm{M_{L}}$ and $\mathrm{M_{R}}$ induces a change in direction for the insect, while similar activities set it on a straight path, autonomously regulating its course as long as the motor neurons remain in control.
As the home vector shrinks, the firing rates of \gls{cpu4} are brought back into equilibrium.

In some instances, the insect first approaches an orthogonal axis (e.g the positive vertical axis in \cref{fig:spike_rates}) before taking a straight line back to its home.
This artifact becomes more pronounced as the maximum traveled distance increases. 
It occurs when the input stimuli of \gls{cpu4} cells become high enough to saturate their output activities towards the maximum spike rate.
As this happens, the home vector, as communicated to \gls{cpu1}, gets compressed in the most dominant direction and therefore rotationally distorted, resulting in the observed behavior.
However, as the distortion only emerges in the output activities of \gls{cpu4}, not in the stored values, it does not influence agent's accuracy or reliability. 

After reaching its home location at $t \approx \SI{100}{\milli\second}$ the insect begins swarming around it for another $\SI{100}{\milli\second}$.
The motor neurons, under the influence of \gls{cpu1}'s alternating decisions, trigger frequent turns, thus maintaining the agent in close proximity.
As it thereby repeatedly approaches and departs from the origin, the home vector in \gls{cpu4} dynamically contracts, expands, and rotates. 

The results, as depicted in \cref{fig:spike_rates}, validate that path integration-based navigation can be realized with accelerated neuromorphic circuits.
They also show that the recorded movements and neural activities closely resemble those observed in biology~\citep{turner2016insect,stone2017anatomically}.
The experiment demonstrates that neural structures derived from insect brains can be mapped onto neuromorphic hardware to effectively control a virtual agent, thereby realizing a biomimetic cybernetic steering system.

\subsection{Statistical performance and optimization}
\begin{figure}
    \resizebox{\linewidth}{!}{\input{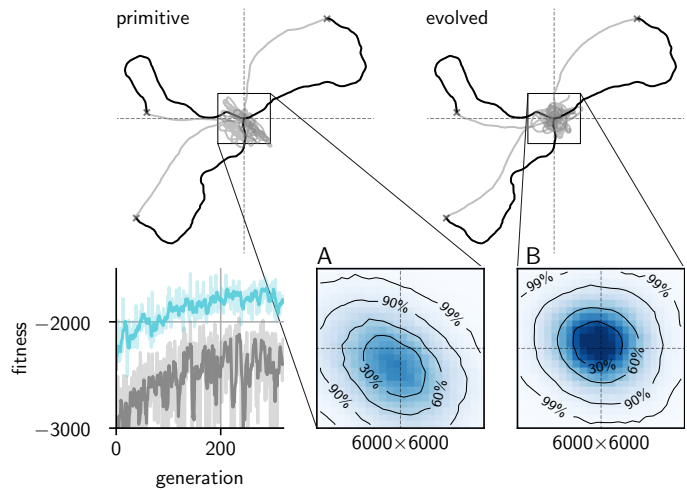}}
    \caption{
        Statistical performance of the evolutionary optimization.
        Three sample trajectories generated by the primitive
        (left) and the evolved network (right).
        Below is a $6000\times 6000$ zoom into the center region that shows a histogram over the data points of the looping phase of 1000 trajectories.
        The primitive network's center of looping is shifted to the lower left with a broad and elliptically deformed looping area.
        The evolved network is more centered and exhibits tighter and more symmetric looping behavior.
        The lower left plot shows the population fitness (gray) and the fitness of the best three individuals (blue).
        For each individual, the faint line is the actual fitness whereas the strong line provides a moving average for better visibility.
        The optimization converges after $\sim$ 200 steps.
    }
    \label{fig:evolution}
\end{figure}

Furthermore, we performed statistical analyses to explore the properties and performance of the emulated insect.
A series of \SI{1000}{} independent experiments reveals that the agent always returns into the correct direction and typically stops at the correct distance.
On average, the return location (computed as the mean position over $\SI{100}{\milli\second}$ during looping) deviates from the actual home location by \SI{842}{\steps}.
This falls within a circle around the origin, accounting for \SI{4.9}{\percent} of the median traveled outbound distance, or merely \SI{0.24}{\percent} of the area covered in by the median outbound journeys.
The histogram over the trajectories during looping, i.e. after $t = 2\cdot t_{\mathrm{return}}$, is shown in \cref{fig:evolution}~A.

The distances given in integer unit lengths can be visualized more intuitively by relating them to physical quantities.
Assuming typical traveling velocities of honeybees in the field (roughly \SI{7}{\meter\per\second} according to~\citep{wenner1963flight}), the median journey radius corresponds to \SI{618}{\meter} (\SI{17309}{\steps}) and the average deviation from the origin to approximately \SI{23.6}{\meter}.
Please note, however, that this might differ substantially when more accurate data for \textit{Megalopta genalis} or other species is taken into account, or when the circumstances are modelled more accurately (considering varying flight speeds for outbound and return trips, obstacles, wind, etc.).
The metric distances should therefore be understood solely as an illustrative aid. 

To further improve the performance of the neuromorphic agent we optimized the synaptic weights based on an evolutionary strategy~\citep{hansen1996adapting}.
The optimization yielded a almost five-fold improvement in the average deviation of the return location, reducing it from \SI{4.9}{\percent} to \SI{1.0}{\percent}.
Using the traveling velocities given above, this corresponds to a reduction from \SI{23.6}{\meter} to \SI{4.7}{\meter}.
We also found that the insect's motion around the return location becomes more narrow, reducing the average looping radius by \SI{22}{\percent}.
The results are listed in more detail in \cref{tab:optimization} and illustrated in \cref{fig:evolution}.

In line with a number of previous studies and experiments~\citep{schmitt2017neuromorphic,billaudelle2021structural,cramer2022surrogate,pehle2022brainscales2,goeltz2021fastanddeep,wunderlich2019demonstratingstudy},
this confirms that device mismatches and inaccuracies intrinsic to analog neuromorphic hardware can be compensated by parameter optimization.

\begin{table*}
    \begin{center}
        \begin{tabular}{ l c c c c }
	    \phantom{empty} & $\mu_{x}/\mu_{y}$ & $\sigma_{x}/\sigma_{y}$ & overlapping & within $r = \SI{1000}{}$\\
            \hline
            primitive & -313/-774 & 1419/1398 & \SI{69.3}{\percent} & \SI{91.1}{\percent}\\[1em]
            optimized & -75/150 & 1125/1066 & \SI{88.7}{\percent} & \SI{98.9}{\percent}\\[1em]
            \shortstack[l]{relative\\improvement} & -76\%/-81\% & -21\%/-24\% & +\SI{28}{\percent} & +\SI{8.6}{\percent}\\
        \end{tabular}
    \end{center}
    \caption{Mean displacement and standard deviation before and after evolutionary optimization.
    In the primitive state \SI{69.3}{\percent} of the return trajectories overlap the home location directly and \SI{91.1}{\percent} are within an $r = \SI{1000}{}$ radius around it.
    After evolutionary optimization this increases to \SI{88.7}{\percent} and \SI{98.9}{\percent}, respectively.}
    \label{tab:optimization}
\end{table*}

\section{Discussion}
This paper describes how a neural network model faithfully derived from physiological observations of insect brains can perform reliable and accelerated two-dimensional path integration on analog neuromorphic hardware.
The navigating agent thereby consists of the physically emulated neuromorphic brain connected to a virtual body via mixed-signal sensors and motor units.
All these components are implemented and autonomously executed on a single neuromorphic chip.

To realize biologically viable integrators, we introduced a single-cell spike-based short-term memory mechanism based on axo-axonic synapses. 
This mechanism has not been implemented on neuromorphic hardware before and we are not aware of any studies in the computational neurosciences that have yet employed such a neural construct for similar purposes. 
As it is application-agnostic, it can potentially be utilized in various other scenarios beyond path integration, where flexible and network-interactive spike-based short-term memory cells might play a role. 

Here, these memory cells provide the key elements for implementing the spike-based neuronal navigation circuit. 
Primarily through them, the underlying network model, which initially relied entirely on heuristically described differential equations~\citep{stone2017anatomically}, could be translated into a network of spiking \gls{lif} neurons. 
To our knowledge, this constitutes the first publication of a purely spike-based path integration model.

To comply with the limitations in network size of the \BSS2 prototype, we reduced the angular resolution of the neural compound from eight to four divisions without compromising the navigational functions.
In addition, we omitted a population of \textit{pontine} neurons which map input from the excitatory \gls{cpu4} populations inhibitorily onto \gls{cpu1}~\citep{stone2017anatomically}. 
This is possible because \BSS2 does not obey Dale's principle~\citep{strata1999dale}, allowing the neuromorphic \gls{cpu4} populations to both excite and inhibit postsynaptic partners.
Therefore, they can fully absorb the function of the \textit{pontine} populations.
With respect to the source model~\citep{stone2017anatomically}, the number of neurons in the path integration circuit (consisting of \gls{tb1}, \gls{cpu4}, \gls{cpu1}, and optionally the \textit{pontine} population) was, hence, reduced from 56 down to 20.
As the head direction neurons in \gls{tb1} are implemented virtually on \BSS2, and since two additional motor neurons were introduced to offload numerical summations, the total number of neurons on hardware is only 18.
The remarkable functional richness that the insect brain generates with so few neurons could thus be replicated and even further condensed on neuromorphic hardware. 

While the primitive network configuration was already able to successfully navigate, its performance could be improved with an evolution strategy-based optimization acting on the synaptic weights. 
The evolution of 15 individuals, each performing an entire spread-out and homing journey, through 320 generations took only 32 minutes on a single neuromorphic core. 
Had this evolution not been carried out on \BSS2 with its highly accelerated neurons but in the same configuration on a real-time system, it would instead have taken more than 11 days. 
Even on a multi-core real-time system with enough compute resources to fully parallelize all individuals, it would still have taken 17.8 hours to breed 320 consecutive generations.
As each generation depends on its predecessor, this execution time can intrinsically not be further reduced on a real-time emulator.

Moreover, as the emulation speed on \BSS2 does not depend on the network size, the rapid experimental throughput can also be obtained for larger networks, and the longer the simulation time is, the more significantly this advantage manifests. 
Neuroplastic developments in mammals, for example, can extend over many weeks, months or even years. 
In order to simulate such processes and iteratively adapt their model parameters to biology, accelerated emulators are, therefore, absolutely indispensable. 

In scenarios where the accelerated neural network needs to interact with external entities, like sensor organs, motor units, the entire body, or its environment, those entities have to be simulated at a higher speed as well.
Otherwise, the various components would act on different time scales and the biological phenomena could, hence, not be reproduced with plausibility and accuracy. 
For the presented agent, the entities belonging to the extraneural domain were therefore modelled and implemented with the same acceleration factor on the digital on-chip co-processor to guarantee consistency and synchronicity between the brain and the body. 

Thus, we demonstrated that the 1000$\times$ acceleration factor enables biologically detailed neurorobotic experiments at an unprecedented throughput, cutting down emulation times by three orders of magnitude as compared to similar studies carried out in biological real-time.
Since the extent of such experiments is not physically limited and the technical capabilities of \BSS2 are by no means exhausted, we are confident that further neurorobotic closed-loop experiments will take advantage of the considerable benefits of accelerated neuromorphic hardware in the future. 
This kind of experiments can substantially contribute to a better understanding of the functional interaction between neuronal structures and their environments.

\section{Methods}
\subsection{Sensory inputs}
\label{sec:methods_inputs}
The neural network receives input from two classes of sensors, the \gls{tb1} compass and the \gls{tn} flow field sensors.
Both use rate encoding to translate the sensory input into spike trains that can be processed by the \gls{lif} neurons at a maximum rate of $r_{\mathrm{max}} = \SI{100}{\kilo\Hz}$ corresponding to $\SI{100}{\Hz}$ in the biological time equivalent.

The \gls{tb1} population comprises four neurons $\mathrm{TB1}_{j}$ with $j = \{0, 1, 2, 3\}$ that encode the head orientation $\phi$ in their output activity:
\begin{equation}\label{equ:tb1}
  r_{\mathrm{TB1}_{j}} = \frac{r_{\mathrm{max}}}{2}\cdot\left[ 1 + \sin\left(\phi + \frac{j\cdot\pi}{2}\right) \right].
\end{equation}

\begin{figure}
    \def\svgwidth{\linewidth}
    {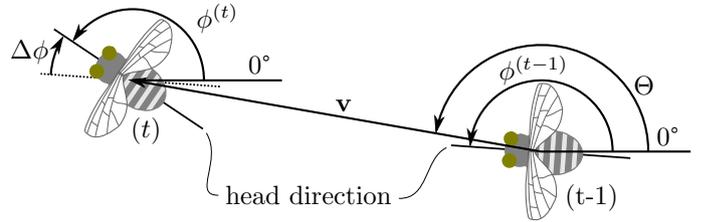}
    \caption{Coordinate system of the insectoid agent.
    $\Theta$ is the direction of flight and $\phi^{(t)}$ and $\phi^{(t-1)}$ the head direction of the current and the previous time step, respectively.
    $\Delta\phi$ is the difference between both and indicates the head rotation.
    As the time interval is discrete and defined to be 1, the traveled distance is equal to the velocity vector $\bm{v}$.}
    \label{fig:agent_coordinates}
\end{figure}

One \gls{tn} flow field sensor measures a scalar velocity into one selected direction of its own movement relative to the environment.
Its output depends on the head rotation $\phi$ relative to the insect's velocity $\bm{v}$, the static rotational separation $\phi_{\mathrm{TN}}$ of the bilateral sensor relative to $\phi$, and the head rotation $\Delta\phi^{(t)} = \phi^{(t)} - \phi^{(t - 1)}$ between the last and the current time step (see \cref{fig:agent_coordinates}):
\begin{equation}\label{eq:flow_field_holonomic}
    r_{\mathrm{TN_{L/R}}} = \mp|\bm{v}|\cdot\sin(\Theta - \phi \pm \phi_{TN}) \pm \rho\cdot\Delta\phi.
\end{equation}
$\Theta$ is the direction of the insect's velocity
\begin{equation}\label{eq:velocity}
    \bm{v} = v \cdot \begin{bmatrix} \cos(\Theta) \\ \sin(\Theta) \end{bmatrix}
\end{equation}
and $\rho$ is a constant scaling factor for the head rotation that is proportional to the interocular distance between the two flow field sensors.
For the experiment, we select $\rho = 2$ and restrict ourselves to $\Theta = \phi$.
\citet{stone2017anatomically} demonstrated that the model can tolerate a significant constant offset $\Theta\neq\phi$.

\Cref{eq:flow_field_holonomic} can therefore be reduced to
\begin{equation}\label{eq:flow_field}
    r_{\mathrm{TN_{L/R}}} = |\bm{v}|\cdot\sin(\phi_{TN}) \pm \rho\cdot\Delta\phi.
\end{equation}

\subsection{Motor outputs}
Following \citeauthor{stone2017anatomically}, the average activity of the left and right \gls{cpu1} populations determines the insect's direction of movement.
We connect all neurons of a $\mathrm{CPU1_{L/R}}$ population to one motor neuron $\mathrm{M_{L/R}}$ in order to compute the average.
On \BSS2 these connections are realized by a single synapse that receives input from all four \gls{cpu1} neurons per hemisphere.

The motor neuron's spike rate is then translated into a directional velocity update
\begin{equation}
    \Delta\phi = \Delta\Theta = \mu\cdot(r_{\mathrm{M_{L}}} - r_{\mathrm{M_{R}}})
\end{equation}
where $\mu = 1.6$ is a heuristically set scaling constant.
The absolute velocity $|\bm{v}|$ stays constant.

\subsection{Neuron model}
\label{sec:neuron_model}
\begin{figure}
    \begin{subfigure}[t]{0.65\linewidth}
        A\newline
        \def\svgwidth{\linewidth}
        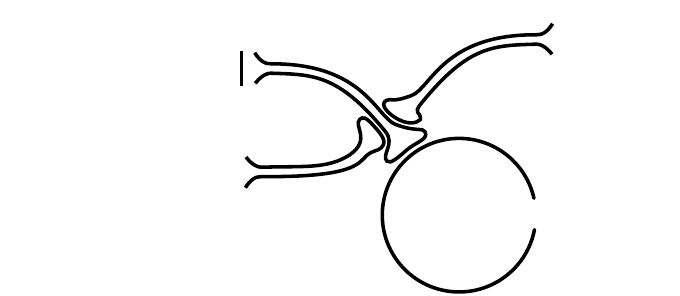
    \end{subfigure}%
    \begin{subfigure}[t]{0.35\linewidth}
        B\\[0.5em]
        \resizebox{\linewidth}{!}{
        \input{figures/py/plots/cpu4_response.pgf}
        }
    \end{subfigure}
    \caption{Axo-axonic depression and facilitation.
    A) A synapse that forms an excitatory connection to a background source with a constant spike rate is modulated by two axo-axonic synapses.
    Activity coming from the $\mathrm{TN_{L/R}}$ axons facilitates the synaptic weight $w_{\mathrm{CPU4}, j}$, while activity from $\mathrm{TB1_{i}}$ inhibits it.
    The neuron responds with an output rate $r_{\mathrm{CPU4}, j}$.
    B) Output rate vs. weight.
    The plot shows an overlay of the transfer functions of eight \gls{cpu4} neurons on \BSS2.}
    \label{fig:axo-axonic}
\end{figure}

The circuit proposed in \citep{stone2017anatomically} is based on a firing-rate neuron model with a synaptic input current
\begin{equation}\label{eq:internal_state_stone}
  I_{j} = \sum\limits_{i} w_{ij}r_{i}
\end{equation}
of the $j^{th}$ neuron.
The output of the $j^{th}$ neuron is obtained by applying a logistic function
\begin{equation}\label{eq:transfer_function_stone}
  r_{j} = \frac{1}{1 + e^{a_{j}\cdot I_{j} - b_{j}}}
\end{equation}
to the internal state $I_{j}$, where $a_{j}$ and $b_{j}$ are two real-valued model parameters that are similar for all neurons.
This transfer function is a commonly used choice to approximate the rate output of spiking neuron models under stochastic influence \citep{stevens1996integrate,brunel1998firing,dayan2001theoretical,fourcaud2002dynamics,moreno2004role,petrovici2016stochastic}.
Our empirical measurements of the analog neurons on the chip have revealed similar transfer curves (see \cref{fig:axo-axonic}B).

\Cref{eq:internal_state_stone,eq:transfer_function_stone} apply to the neurons in the three subcircuits \gls{tn}, \gls{tb1}, and \gls{cpu1}. 
The neurons in \gls{cpu4}, which are responsible for integrating over the traveled path, behave differently.
Their internal state is given by
\begin{equation}\label{eq:cpu4_internal_state}
  I^{(t)}_{\mathrm{CPU4}} = I^{(t-1)}_{\mathrm{CPU4}} + h\cdot\left( r^{(t)}_{\mathrm{TN}} - r^{(t)}_{\mathrm{TB1}} - k \right)
\end{equation}
with a discrete time $t$, coupling strength $h$ and decay constant $k$.
It therefore depends on its own previous internal state and the output rates of \gls{tn} on \gls{tb1}. 
The time constant of this dynamical process is roughly on the order of the duration of the entire journey, i.e. around \SI{200}{\second} in biological time.

Since the synaptic time constant of typical neurons is on the order of \SIrange{1}{100}{\milli\second} and the membrane potential is reset after each firing event, \cref{eq:cpu4_internal_state} cannot not solely emerge from the dynamics of a single \gls{lif} unit. 
Also, a single-cell feedback mechanism based on self-recurrent synapses would not allow for an integration over these time scales while maintaining the necessary precision and stability. 
Since each \textit{glomerulus}~\citep{turner2016insect} in the respective compartments of the \textit{central complex} contains exactly one \gls{cpu4} neuron and only a small number of other neurons ($\mathcal{O}(\SI{1}{} - \SI{10}{})$), it is also unlikely that the time scales required by \cref{eq:cpu4_internal_state} arise from recurrent connections within small clusters of \gls{lif}-like cells.

Neural adaptations (as modeled by AdEx-equations~\citep{brette2005adaptive} for instance) neither act on suitable time scales, as they typically last for no more than several seconds (usually several hundred ms)~\citep{naud2008firing}.
The same holds true for short-term synaptic facilitation and depression~\citep{tsodyks1998neural}.

Among the remaining options are long-term synaptic plasticity and synaptic modulations, in particular, presynaptic facilitation and depression~\citep{byrne1996presynaptic, lodish2012molecular}.
As described in \cref{sec:results_network}, this type of synaptic modulation involves axo-axonic synapses which are formed between the axons of facilitating or depressing presynaptic neurons and axon terminals of postsynaptic neurons. 
While the axo-axonic synapse itself remains constant over time, the activity it projects on the postsynaptic axon terminal modulates the weight of the following synapse between the targeted terminal and the corresponding postsynaptic neuron. 
Presynaptic modulations can occur on short-term memory time scales ($\mathcal{O}(\SI{10}{\minute})$)~\citep{byrne1996presynaptic} and, therefore, comply with the demands of \cref{eq:cpu4_internal_state}.

We implement \cref{eq:cpu4_internal_state} through axo-axonic weight modulations of excitatory synapses coming from background spike sources that are firing periodically at a constant rate (see \cref{fig:axo-axonic}).
The strengths of those synapses are reduced or increased on short-term memory time scales by presynaptic facilitation or depression~\citep{byrne1996presynaptic, lodish2012molecular}.
In our case, these neurons correspond to the \gls{tn} and \gls{tb1} subpopulations.
Thus, in order to reproduce the behavior of \cref{eq:cpu4_internal_state} on \BSS2, one background spike source with a constant average rate is connected to each integrating neuron through an excitatory synapse with a variable weight $w_{\mathrm{CPU4},j}$.
The integrating neuron therefore responds with an average output spike rate $r$ that is proportional to this weight.
To that end, the synapse realizes the short-term memory dynamics which are necessary for the distance integration.
In analogy to \cref{eq:cpu4_internal_state}, we have
\begin{equation}\label{eq:weight_update}
  w^{(t)}_{\mathrm{CPU4}} = w^{(t-1)}_{\mathrm{CPU4}} + h\cdot\left( r^{(t)}_{\mathrm{TN}} - r^{(t)}_{\mathrm{TB1}} - k \right)
\end{equation}
A parameter sweep revealed an optimal parameter combination of $k = 2$ and $h = 0.0336$.

We would like to add that the background spike sources fire periodically due to the tight timing demands of the code execution on the \BSS2 prototype.
Spike sources with Poisson-distributed numbers of spikes per time interval would be favorable because they would introduce a more stochastic behavior, blurring out artifacts, and appear biologically more plausible.
However, in simulation we found that the relevant statistical performance is not measurably affected by the strictly periodic sources and in hardware, jitter and intrinsic analog noise additionally mitigate possibly introduced artifacts to some degree.
Moreover, the current full-scale version of \BSS2 offers configurable on-chip Poisson spike generators and, hence, renders the circumvention by periodic sources in future experiments obsolete.

\subsection{Experiment scheduling}
In order to run the environment and agent simulation in temporal coherence with the dynamics of the neural network, code execution has to be carefully scheduled.
Every experimental run starts with transferring a set of parameters from the host computer to the \BSS2 system.
This includes a random seed, the experiment run time $t_{\mathrm{stop}}$, the time at which the agent is to return $t_{\mathrm{return}}$, the \gls{cpu4} decay $k$, the \gls{cpu4} update scaling $h$, and the weights of all but the \gls{cpu4} input synapses.
Subsequently, the code execution is started on the \gls{ppu}.

During runtime, the \gls{ppu} executes the simulation and implements the sensory input spike sources.
One agent/environment update is conducted every $\Delta t = \SI{100}{\micro\second}$, which corresponds to a movement update rate of \SI{10}{\Hz} in the biological time equivalent.
To achieve a maximum biological spike rate of \SI{100}{\Hz}, i.e., \SI{100}{\kilo\Hz} on the chip, the spike sources have to be able to send a spike every \SI{10}{\micro\second}.
As the code execution time for one full agent/environment update exceeds the time between two consecutive spikes, it is divided into three subcycles that are called between spike sending routines:
\begin{itemize}
    \item iterate the insect state and transmit the sensory input to the \gls{tn} and \gls{tb1} spike generators,
    \item read the \gls{cpu4} weights and calculate their updates,
    \item write back the updated \gls{cpu4} weights, process the motor neuron output and update the agent velocity accordingly, update the simulation state, and store the trajectory data.
\end{itemize}

At $t = t_{\mathrm{return}}$, the agent switches from random to model-driven movement.
The simulation stops at $t = t_{\mathrm{stop}}$.
At that point, the trajectory data and spikes are read back from the host computer.

\begin{figure}
    \includegraphics[width=\linewidth]{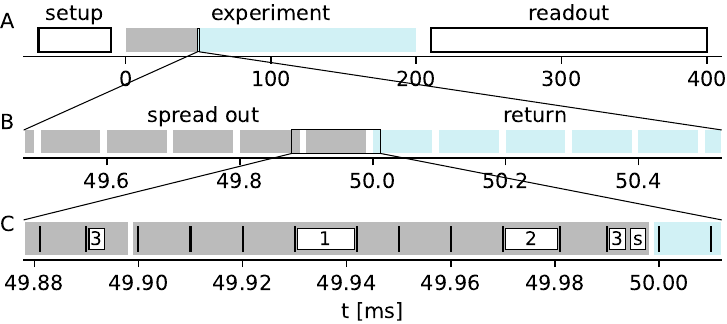}
    \caption{Experiment schedule.
    A) The experiment starts with a setup phase in which the parameters are transferred to the \BSS2 prototype and the synaptic weights are initialized.
    The actual experiment, starting with a spread out phase follows.
    After that, data is read back from the system memory to the host.
    This phase's execution time is stochastic and depends on the amount of spikes produced in an experiment.
    B) Zoom into the point of return.
    Each block symbolizes an agent/environment update, the blue color indicates the return phase.
    C) Zoom into one update.
    The black bars mark the times at which the spike sending routine is called.
    The three update phases are represented by the white blocks \textit{1}, \textit{2}, and \textit{3}.
    Position recording happens only in every second update cycle in block \textit{s}.
    Note that the variable duration of the update phase executions can introduce jitter of the following spike.
    }
    \label{fig:schedule}
\end{figure}

The \BSS2 prototype chip can record all post-synaptic spikes using the connected FPGA, while the available on-chip memory for arbitrary data storage is limited to \SI{4}{\kilo\byte}.
The latter is used for saving the trajectory from which the velocity and the optical flow can be reconstructed.
Both the x and y coordinate are \SI{16}{\bit} signed integer values.
Therefore, a total of 1000 locations can be stored.
With $n = t_{\mathrm{stop}}/\Delta t = 2000$ time steps in one run, every second step is written to memory.
\Cref{fig:schedule} gives an overview of the scheduling.

\subsection{Discretization of synaptic weights}
While in \citep{stone2017anatomically}, the internal state of the integrator neurons have floating point precision and a corresponding dynamic range, the  \SI{6}{\bit} discretization on \BSS2 limits the number of possible weights per synapse to 64.
In order to extend this range, we additively join 16 synapses together into one supersynapse.
One presynaptic partner is connected to one postsynaptic neuron over 16 synapses, forming one supersynapse with a possible weight of $0$ to $16\cdot 63 = 1008$.
The dynamic range is therefore extended from \SI{6}{\bit} to almost \SI{10}{\bit}.

\subsection{Calibration}
Due to their analog design, the neuronal circuits on \BSS2 are subject to temporal and fixed-pattern noise.
The latter is caused by manufacturing variations and results in instance-to-instance variations of the neuron and synapse parameters.
In order to mitigate this effect, we calibrate the neuronal firing rates on each chip and store the results in a static data bank.
For this experiment, two calibration routines are particularly important:

The \gls{cpu1} and motor neurons are calibrated such that their output rate is proportional to $r_{\mathrm{out}} = r_{\mathrm{exc}}\cdot (1 - r_{\mathrm{inh}})$, where $r_{\mathrm{exc}}$ and $r_{\mathrm{inh}}$ is the excitatory and inhibitory input rate, respectively.

The \gls{cpu4} neurons, on the other hand, are calibrated to respond with an output rate that is proportional to the weight $w_{\mathrm{CPU4}, j}$ (see \cref{fig:axo-axonic}B).
All rates are normalized to a maximum of $r_{max} = \SI{100}{\kilo\Hz}$ and a minimum of $r_{min} = \SI{0}{\Hz}$.
Functionally, it is relevant that the neuron parameters match well around the operating point $r = 0.5\cdot r_{max}$, while the behavior at the range margins is of minor importance.

\subsection{Evolutionary optimization}
To mitigate fixed-pattern noise effects and further optimize network performance, we employ an evolution strategy with covariance matrix adaptation~\citep{hansen1996adapting}.
The optimization parameters are the 26 synaptic weights $\bm{w}$ that are connected to the \gls{cpu1} population: \gls{tb1}~$\rightarrow$~CPU1, CPU4~$\rightarrow$~CPU1 (inhibitory and excitatory), and CPU1~$\rightarrow$~M.
To optimize for tight and precise looping around the home position, the fitness $f_{i}$ of an individual $i$ is derived from its trajectory $\bm{x}_{i}(t)_{t = [2\cdot t_{\mathrm{return}}, t_{\mathrm{stop}})}$ during looping:
\begin{equation}
    f_{i} = -\left\langle \left| \left\langle \bm{x}_{i} \right\rangle_{t} \right| + \left| \sqrt{\left\langle (\bm{x}_{i} - \langle \bm{x}_{i} \rangle)^2 \right\rangle}_{t} \right| \right\rangle_{\mathrm{runs}}
\end{equation}
The first term is the time-averaged radial distance to the home location and the second term the time-averaged looping diameter.
This sum is then additionally averaged over three runs with different outbound trajectories.

Instead of selecting a fixed number of best individuals of each generation, we take a weighted sum over the parameters of all individuals to derive the new mean parameter vector for the offspring population:
\begin{equation}
    \bm{\mu} = \sum\limits_{i} p_{i}\bm{w}_{i}
\end{equation}
$p_{i}$ is obtained as
\begin{equation}
    p_{i} = \frac{\tilde{p}_{i}}{\sum\limits_{i}{\tilde{p}_{i}}};\; \tilde{p}_{i} = f_{i}^{-8}
\end{equation}
The power of eight increases the impact of high-fitness genomes on the seed genome for the next generation, while the impact of low-fitness genomes is scaled down. 
Thus, weak individuals are not completely sorted out but still contribute to the next generation to a low degree. 
The exponent therefore implements a soft selection. 
Note that $p_{i}$ is positive due to the even exponential index.
Moreover, small absolute values of $f_{i}$ translate to a high $p_{i}$ and vice versa.

The new population consists of 15 weight vectors drawn from a multivariate Gaussian distribution:
\begin{equation}
    \bm{w}_{i} \sim \mathcal{N}(\bm{\mu}, \sigma\cdot\bm{\Sigma})
\end{equation}
Here, $\sigma = 0.3$ is a heuristically chosen step size and $\bm{\Sigma}$ is a covariance matrix derived from the previous generation
\begin{equation}
    \bm{\Sigma} = \sum\limits_{i}{p_{i}\bm{d}_{i}\otimes\bm{d}_{i}}
\end{equation}
with $\bm{d}_{i} = \bm{w}_{i} - \bm{\mu}$.
In this way, the variance is increased into the direction of successful mutations.
Typically, the optimization converges after approximately 200 generations.

\section*{Author Contributions}
KS conceived the project and conceptualized, implemented and evaluated all experiments.
TW implemented a simulation of the experiment, verifying its feasability.
TW, EM, PS, and YS contributed to the development of the \BSS2 operating system.
PS offered consulting and support concerning the \gls{ppu} compiler.
SB, BC and YS developed parts of the calibration routines used in this work.
CP implemented the \gls{fpga}-based spike router.
MAP contributed to the manuscript and to the development of the \BSS2 system.
JS is the leading hardware architect of the \BSS2 system and head of the Electronic Vision(s) research group at Heidelberg University.
Karlheinz Meier was the head of the research group in the early stages of this work and provided the initial financial, infrastructural, and motivational foundation without which none of this research could have been realized.
This manuscript is dedicated to him personally and adds yet another facet to his extensive scientific accomplishments.

\ifarxiv

\else
	\bibliography{bibliography}
\fi

\end{document}